\documentclass[10pt,journal,compsoc]{IEEEtran}
\usepackage{graphicx}
\usepackage{caption}
\usepackage{subcaption}
\usepackage{multirow}
\usepackage{amsmath}
\usepackage{amsfonts}

\def\H{\mathbf{H}}

\def\H{\mathbf{H}}

\def\G{\mathbf{G}}

\def\z{\mathbf{z}}

\def\Z{\mathbf{Z}}
\def\0{\mathbf{0}}

\def\G{{G}}

\def\A{\mathbf{A}}

\ifCLASSOPTIONcompsoc
  \usepackage[nocompress]{cite}
\else
  \usepackage{cite}
\fi
\ifCLASSINFOpdf
\else
\fi
\begin{document}
%
\title{Memory Efficient Temporal \& Visual Graph Model for Unsupervised Video Domain Adaptation}
%
%
%
%

\author{Xinyue~Hu,
        Lin Gu,
        Liangchen Liu,
        Ruijiang Li,
        Chang Su,
        Tatsuya Harada,
        Yingying Zhu
\IEEEcompsocitemizethanks{\IEEEcompsocthanksitem 
Xinyue Hu and Yingying Zhu are with the Department of Computer Science and Engineering, The University of Texas at Arlinton, Arlington,
TX, 76019.\protect\\
E-mail: xxh4034@mavs.uta.edu, yingying.zhu@uta.edu
\IEEEcompsocthanksitem 
Lin Gu and Tatsuya Harada are with The University of Tokyo, Tokyo 113-8654, Japan, and RIKEN, 2-1 Hirosawa, Wako, Saitama, 351-0198, Japan. \protect\\
E-mail: lin.gu@riken.jp, harada@mi.t.u-tokyo.ac.jp
\IEEEcompsocthanksitem 
Liangchen Liu is with National Institutes of Health Clinical Center, Bethesda, MD, United States. \protect\\
E-mail: liangchen.liu@nih.gov
\IEEEcompsocthanksitem 
Ruijiang Li is with eBay Inc. 
E-mail: lithium7456@gmail.com
\IEEEcompsocthanksitem 
Chang Su is with the Department of Health Service Administration and Policy (HSAP), College of Public Health, Temple University.
\protect\\
Email: su.chang@temple.edu
}
}

\IEEEtitleabstractindextext{%
\begin{abstract}
Existing video domain adaption (DA) methods need to store all temporal combinations of video frames or pair the source and target videos, which are memory cost expensive and can't scale up to long videos. 
To address these limitations, we propose a memory-efficient graph-based video DA approach as follows. 
At first our method models each source or target video by a graph: nodes represent video frames and edges represent the temporal or visual similarity relationship between frames.  
We use a graph attention network to learn the weight of individual frames and simultaneously align the source and target video into a domain-invariant graph feature space. 
Instead of storing a large number of sub-videos, our method only constructs one graph with a graph attention mechanism for one video, reducing the memory cost substantially.
The extensive experiments show that, compared with the state-of-art methods, we achieved superior performance while reducing the memory cost significantly. 
\end{abstract}

\begin{IEEEkeywords}
Computer vision, Video analysis.
\end{IEEEkeywords}}

\maketitle

\IEEEdisplaynontitleabstractindextext

%
\IEEEpeerreviewmaketitle

\IEEEraisesectionheading{\section{Introduction}\label{sec:introduction}}

\begin{figure}[ht]
\centering
    \begin{subfigure}[b]{0.45\textwidth}
    \centering
    \includegraphics[width=\textwidth]{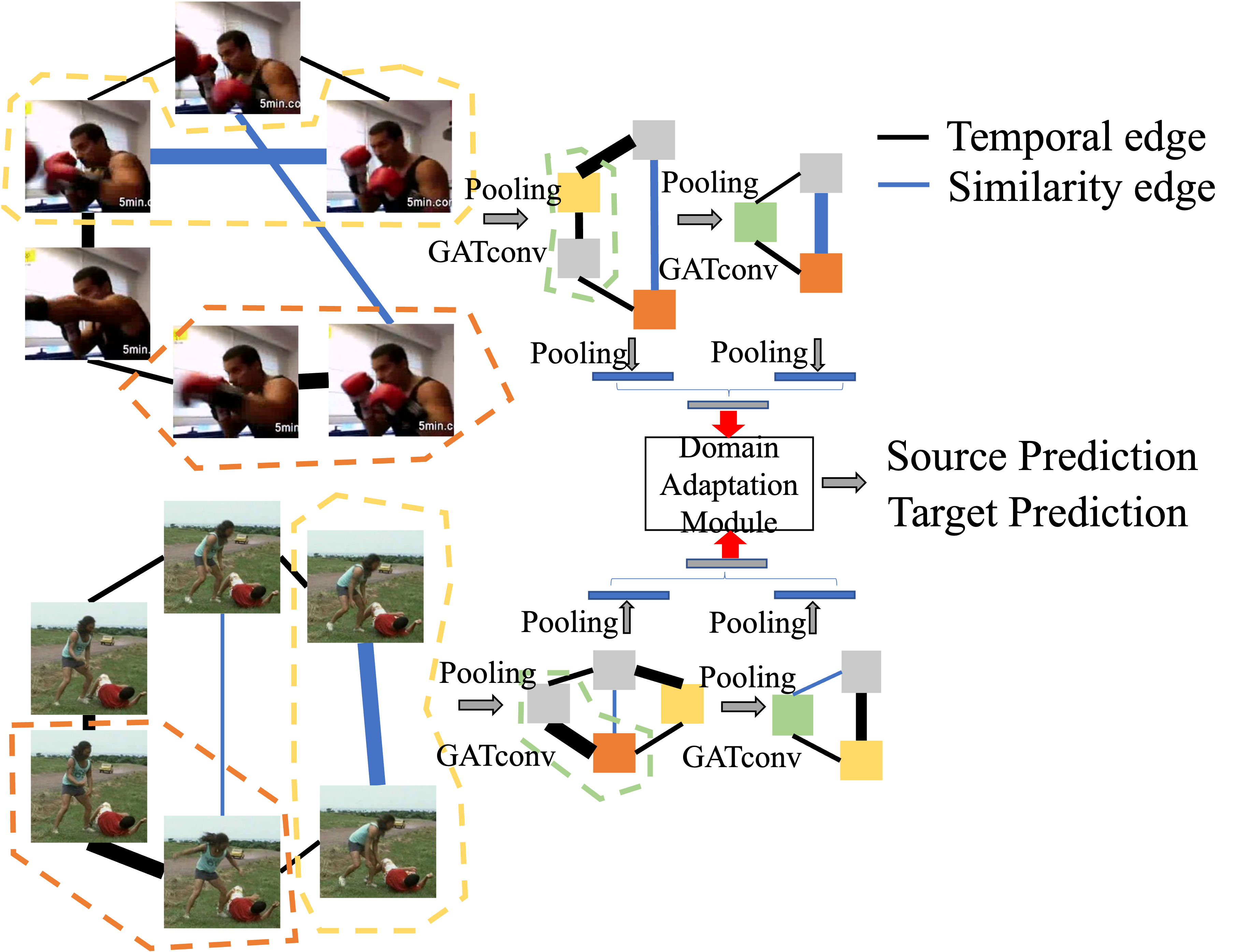}
    \caption{The workflow of our proposed method}
    \label{fig:fig1a}
    \end{subfigure}
    \begin{subfigure}[b]{0.45\textwidth}
    \centering
    \includegraphics[width=\textwidth]{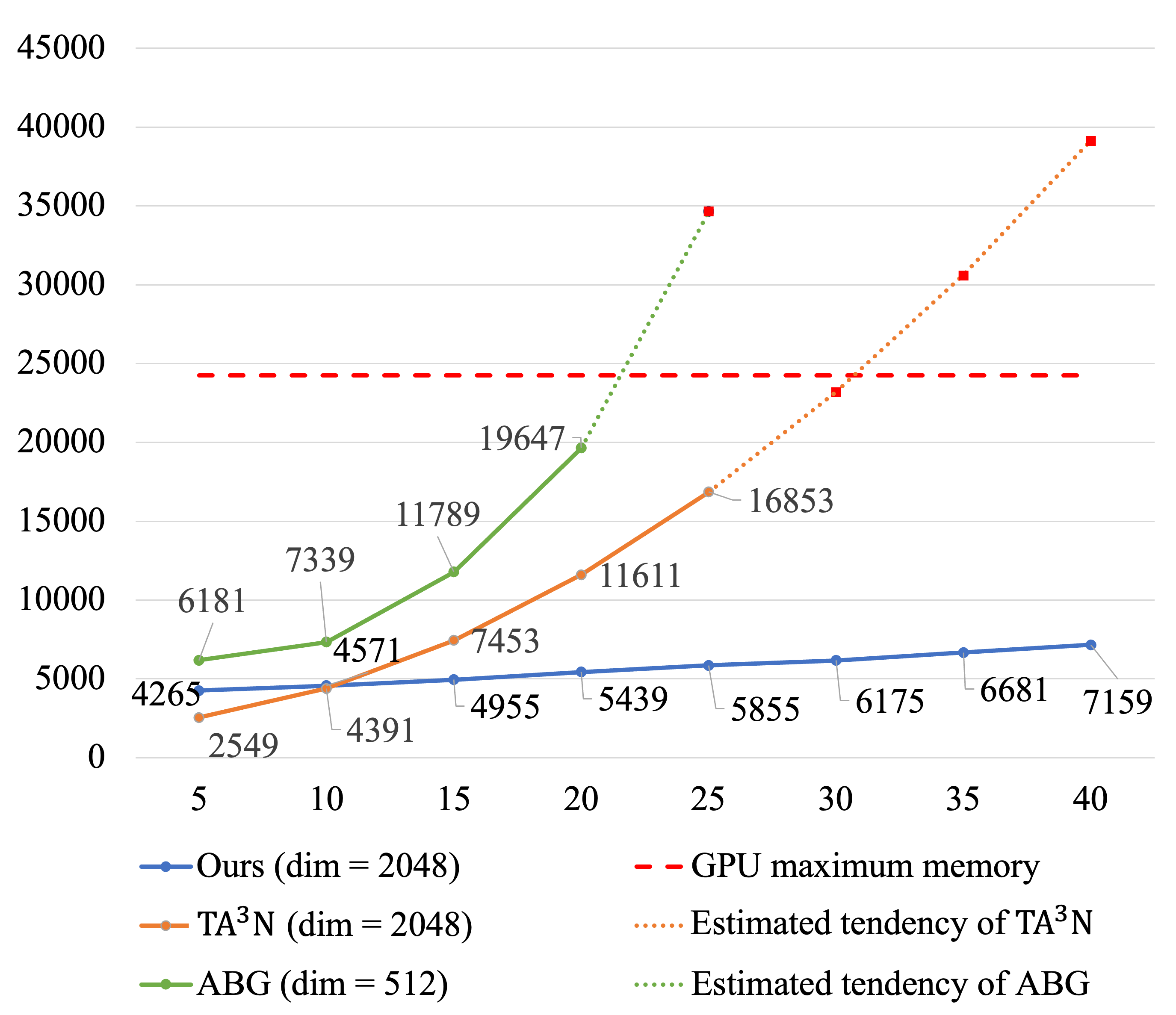}
    \caption{Comparison of GPU memory usages of TA$^3$N, ABG, and our method}
    \label{fig:fig1b}
    \end{subfigure}
\caption{Our method includes three major contributions: 1) constructing a graph model to present source and target video frame features; the graph edge is constructed based on temporal relationship and visual feature relationship; 2) learning a domain-invariant feature space for both source and target video graphs (GATconv represents Graph Attention Network convolution); 3) our approach reduced the temporal frame alignment cost significantly compared to the temporal attention approach
Temporal Attentive Adversarial Adaptation Network (TA$^3$N)~\cite{videoDAICCV} and Adversarial Bipartite Graph Learning (ABG)~\cite{ABG}.
}
\label{fig:fig1}
\end{figure}

%
%
%
%
\IEEEPARstart{D}{ue} to the recent explosion of online video applications, video content analysis has emerged as an important task. However, its success relies on large-scale labeled training data whose collection is time-consuming and cost expensive. 
In the meantime, if the training data differs from the testing data, the video analysis algorithms would suffer a performance drop. As shown in Figure~\ref{fig:fig1a}, a model trained in the gym scenario would degrade significantly when applied on outdoor one, even though both videos are about boxing.  

Unsupervised Domain Adaptation (UDA) is an effective solution for the data distribution shift problem.
Despite its prevalence in image analysis, little effort was spent on video UDA ~\cite{videoDAICCV,sava_eccv20,ABG}. One of the challenges in video UDA lies in the distribution shift of both image-level (similar to image DA) and temporal frame-level between source and target domains. 
For example, the starting and ending time of the punching and the environment (indoor vs outdoor) differs in source and target domains in Figure~\ref{fig:fig1a}. 
Current video UDA addressed it by either using temporal frame attention mechanisms~\cite{videoDAICCV,sava_eccv20} or constructing a temporal source and target frame-level aligning graph~\cite{ABG}.  
The temporal frame attention mechanism constructs a set of sub-videos by extracting all temporal combinations of target-video frames to handle the different starting/ending times and applies temporal attention to these sub-videos. 
This strategy requires a large training memory to store the sub-videos, which would jeopardize its scalability to long videos with large frame number~\cite{sava_eccv20,videoDAICCV}
($O(nm^2v)$-Table\ref{tab:cost}).
One recent work proposed to leverage contrastive learning on learning the domain-invariant video feature representations~\cite{song2021spatio}, however, a large number of negative sub-videos are needed to be stored in memory($O(nm^2v)$-Table~\ref{tab:cost}).
Thus, it is extremely computationally expensive.
To avoid sub-videos extraction, an alternative solution is to pair similar source and target videos before learning a bipartite graph to match the source and target frames for video DA~\cite{ABG}.  
Nevertheless, pairing the source and target videos is also an NP-hard problem with high computational cost.  
Its computational cost will be dramatically increased with the number of videos in the dataset.($O(n^2v^2)$-Table~\ref{tab:cost})

To reduce the large memory and computational requirement in video DA, we propose to construct a novel memory efficient graph with attentional convolution to represent each source or target video and learn a domain-invariant graph feature space for DA. 
Figure~\ref{fig:fig1a} shows two videos of ``boxing'' at the gym or in the park with different temporal starting and ending frames related to the activity ``boxing''. 
We represent each frame as a graph node and connect the nodes via temporally adjacent edge or visually similar edge for these two videos. 
The temporally adjacent edge preserves the short-term frame-level relationship and visually similar edge models the long-term frame-level relationship. 
By leveraging two types of edges, our graph model can model both short and long-term frame-level relations to learn the domain-invariant video feature representations. 
We use a three-hidden layer graph convolutional networks with graph attention and pooling layers to extract the graph-based feature representations of the source and target videos. 
These graph-based features are fed into a domain adaptation module to learn a domain-invariant video-level graph feature space. 
These models are trained in an end-to-end framework. 
The learned graph attention weights indicate the importance (highlighted by edge-width) of video frames, and the graph pooling layer can extract sub-graphs which can be considered equally to sub-videos. To this end, our method has similar outcomes as the temporal attention mechanisms with sub-videos.
Instead of storing multiple sub-videos with large memory costs, we only construct one graph to model one video, therefore, our method reduces the memory cost substantially. 
($O(nv)$-Table~\ref{tab:cost})
We show our model's memory cost in Figure~\ref{fig:fig1b} compared to temporal attention-based method for sub-videos~\cite{videoDAICCV} (TA$^3$N ) and bipartite graph-based method for both frames and sub-videos~\cite{ABG}. Our memory cost is slightly increasing with the number of frames, while the memory cost of TA$^3$N and ABG increases very rapidly.
In addition, unlike the bipartite graph-based approach which required pairing source and target videos as input, 
our method takes the unpaired source and target videos directly. Thus, it can be easily applied to large-scale video datasets. 
Our contributions can be summarized as follows:
\begin{itemize}
    \item We propose a novel memory-efficient video UDA method, by representing the source and target videos as graphs with graph attention networks. We learn a domain-invariant graph-based feature space for video UDA.
    \item We propose a generic model which could be integrated into popular DA backbones. We demonstrate the implementation on Domain Adversarial Neural Network (DANN) \cite{ganin2015unsupervised} and Conditional Adversarial Domain Adaptation (CDAN)~\cite{long2018conditional} and show the improved performance compared to the state-of-the-art video DA methods.  It is worth noting that our approach can be easily adapted to any different domain adaptation backbones. We only chose two representative DA methods in this work to demonstrate the effectiveness of our approach. 
    \item  In the experiments, our method shows better performance compared to all state-of-the-art approaches (expensive memory cost) with light memory and computational cost. Our model significantly reduces about 65\% GPU memory size compared to TA$^3$N on 25 frames. 
\end{itemize}

\section{Related Work}
\subsection{Domain Adaptation}
Unsupervised domain adaptation aims at adapting a model trained on labelled source domains to unlabelled target domains. 
One common idea is to learn a shared domain-invariant feature space ~\cite{long2015learning} on both source and target data. 
Different metrics have been proposed to measure the feature distribution difference between two different domains, including Maximum Mean Discrepancy (MMD) \cite{long2015learning}, matching distribution statistical moments at different orders \cite{sun2016deep}, minimum global transportation cost \cite{bhushan2018deepjdot},
adversarial learning~\cite{ganin2015unsupervised} and conditional adversarial learning \cite{long2018conditional}. These DA methods focus on the cross-domain classification problem on images. 

Recent work \cite{videoDAICCV} proposed a temporal attentive model for cross-domain video analysis and their model's training memory cost is enormous for the size of frames. 
\cite{sava_eccv20} studied spatial attentive feature learning for video domain adaptation. 
In \cite{Pan2020AdversarialCA}, the authors proposed a co-attention model on spatial and temporal video segments with impressive results for cross-domain activity recognition. 
However, their spatial-temporal co-attention model uses temporal video segments without considering the temporal sequence alignment problem. 
Although there are some temporal sequence domain adaptations in the image text recognition problem \cite{Zhang2019SequenceToSequenceDA}, none of them jointly considers temporal sequence alignment and spatial attentive feature learning for cross-domain video analysis.

\subsection{Temporal Sequence Alignment}
Aligning two different temporal sequences to compute the sequence similarity has been studied for decades. The typical method is to learn a dynamic time warping (DTW) path which matches different time points in sequence data. When learning the temporal dynamic warping, the order of the sequence data needs to be preserved.  Different DTW methods have been proposed using different time point distance metrics including Canonical Correlation Analysis (CCA)~\cite{CDTW} and deep neural networks \cite{DCTW,DTWnet}. The alignment metric is always solved using dynamic programming to preserve the temporal order and reduce the computational cost.
However, one-by-one time points matching is unnecessary in video domain adaptation and is difficult to be applied because of the strict pairing requirement of source and target videos during training.

Recent works on videos domain adaptation didn't mandate frame-by-frame alignment of time sequences.
Instead, they sought attention mechanisms~\cite{videoDAICCV,Pan2020AdversarialCA}, order combinations~\cite{videoDAICCV,sava_eccv20}, aligning graphs~\cite{ABG}, and contrastive learning~\cite{song2021spatio,kim2021learning} to extract optimal temporal features.
Nevertheless, they are still plagued by problems of choosing start/end time and high GPU memory usage.
In our work, we propose to learn a graph attention model to align the target and source frames for domain adaptation without requiring pairwise source/target videos, and with low space complexity.

\section{Methodology}
 \begin{figure*}[th]
\centering
 \includegraphics[width=1\textwidth]{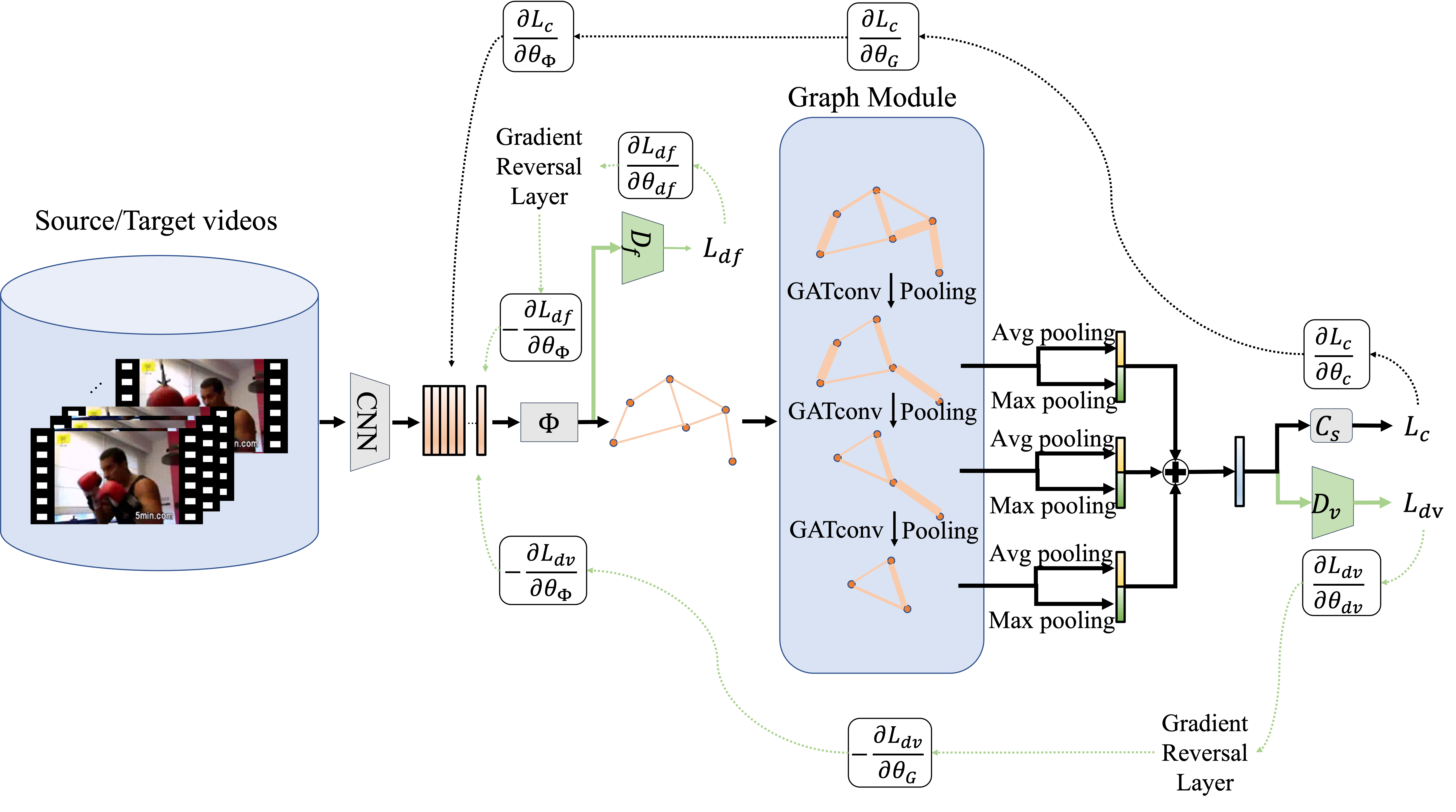}
\caption{Proposed framework on domain adaptive video classification.
We propose 1) construct a graph representation on both source and target videos; 2) learn a graph feature representation for each video by using graph attention and graph pooling (GATconv represents Graph Attention Network convolution); 3) construct a domain-invariant latent graph feature space using different domain adaptation backbones with two level domain discriminator: frame-level and video-level.}
\label{fig:overview}
\end{figure*}

\subsection{Classic Domain Adaptation on Videos}
Given a source video dataset $\mathcal{Z}_s=\{\Z_i^s\}_{i=1}^{n_s}$ with $n_s$ video samples, where a video sample $\Z_i^s$ with its frame-level features can be represented as $\Z_i^s = [\z_{i,1}^s,\cdots,\z_{i,T_i}^s]$,  $T^s_i$ is the number of frames for sample $\mathbf{X}_i^s$ from the source domain $\mathcal{D}_s$ with labels $\mathcal{Y}_s =\{y^s_i\}_{i=1}^{n_s}$, where $y^s_i \in \{1, 2, ..., K\}$ for $K$ different classes.

Similarly, we have an unlabeled target video dataset $\mathcal{Z}_t=\{\Z_i^t\}_{i=1}^{n_t}$, where $\Z_i^t =[\z_{i,1}^t,\cdots,\z_{i,T_i}^t]$ sampled from the target domain $\mathcal{D}_t$ and $T_i^t$ is the video frame number for $\Z_i^t$.
The source domain $\mathcal{D}_s$ has a different distribution from the target domain $\mathcal{D}_t$'s distribution. 
We use the Resnet-101\cite{he2016deep} and ``Two-Stream Inflated 3D ConvNets''(I3D)~\cite{carreira2017quo} using RGB modality as the pre-trained feature extractor to pre-process the video frames.

Most current DA methods assume a shared domain-invariant feature space $\mathcal{Z}$. They aim to learn a shared mapping on the extracted source and target features: $\Phi:\mathcal{Z}_s\to \mathcal{Z}$ and $\Phi:\mathcal{Z}_t \to \mathcal{Z}$ respectively to map into the shared domain-invariant feature space ~$\mathcal{Z}$. 
The same classifier $C:\mathcal{Z} \to \mathcal{Y}$ trained on the labeled source data can be applied to the target data in the shared feature space $\mathcal{Z}$. Both mapping function $\Phi$ and classifier $C_s$ are neural networks with parameters $\theta_{\Phi}$  and $\theta_{C_s}$ respectively.
Without considering the temporal alignment in video domain adaptation, 
the general loss function of cross-domain video classification can be written in two parts as follows:
\begin{equation}
\mathcal{L}_{DA} (\theta_{\Phi},\theta_{C_s}) = \mathcal{L}_{c}(C_s(\Phi(\mathcal{Z}_s)), \mathcal{Y}_s) 
+ \mathcal{L}_d(\Phi(\mathcal{Z}_s), \Phi(\mathcal{Z}_t))
\label{eq:da_loss1}
\end{equation}
where $\mathcal{L}_c$ is the classification loss, aiming to minimize the classification error on source data with labels by the cross-entropy loss~\cite{bishop2006pattern}. 
$\mathcal{L}_d$ is the domain discrepancy loss, minimizing the difference of the feature distribution between source and target data through various metrics such as KL-divergence\cite{tzeng2017adversarial}, MMD distance\cite{ganin2015unsupervised} or
Wasserstein-distance\cite{xu2020reliable,bhushan2018deepjdot}. The above loss function Eq.~\ref{eq:da_loss1} has been successfully applied to plenty of DA problems including image classification \cite{long2018conditional} and text classification \cite{sun-etal-2019-utilizing}.

\subsection{Our Method}
Compared to the classic video domain adaptation models, we propose to investigate and improve the current models from two aspects: 1) model the frame-level features in both source/target videos as graph nodes and construct the graph edges based on the temporal relationship and feature correlation. 2) leverage graph attention to extract video-level feature representation and train a domain adaptation model using both video-level features and frame-level features.

The overall architecture  of our framework is shown in Figure~\ref{fig:overview}. At first, we construct a graph to represent each video with graph adjacent matrix $\A_i^s \in \mathbb{R}^{T_i^s \times T_i^s}$ for source video and $\A_i^t \in \mathbb{R}^{T_i^t \times T_i^t}$ for target video.
We denote {$\mathbf{V}_s^i \in \mathcal{V}_s$} as the set of node feature for source video and {$\mathbf{V}_t^i \in \mathcal{V}_t$} as the set of node feature for target video. 
Each node represents a frame of the video.
We will construct two types of edges as connections, which are temporal adjacent edge connections and similarity edge connections. They can be defined as follows:
\begin{eqnarray}
\nonumber &&\nonumber\A_{mn} = 1, \mbox{\parbox{5cm}{if the $m$-th and $n$-th video frames\\are temporally adjacent,}} \\
&&\nonumber \A_{mn} = 1, \mbox{if } -D(\mathbf{v}_m,\mathbf{v}_n) \mbox{ has the top K score},\\
&&\nonumber \A_{mn} = 0, \mbox{else.}
\end{eqnarray}
where $D(\cdot, \cdot)$ is the Euclidean distance function. {K is the number of similarity edges, which is set to 5.} As shown in Figure~\ref{fig:fig1a}, the black line represents the temporal adjacent edges, the blue line represents the similarity edges. 

We construct a 3-layer graph convolutional neural networks with attention module\cite{velivckovic2017graph} and edge pooling layer\cite{diehl2019edge} to learn the video-level feature representations. 
In Figure~\ref{fig:overview}, the width of the edges expresses the attention weights between the nodes.
The {$l$-th} graph attention convolution layer $\Psi_l$ with edge pooling for the source video can be expressed as follows:
\begin{eqnarray}
\mathbf{V}_s^{l+1},\A_s^{l+1}= \Psi_l (\mathbf{V}_s^l,\A_s^l)
\end{eqnarray}
For the $i$-th layer, the graph representation $\mathbf{h}_i^s$ for source video can be represented as:
\begin{eqnarray}
\mathbf{h}_s^l = [GAP(\mathbf{V}_s^l);GMP(\mathbf{V}_s^l)]
\end{eqnarray}
Where $GAP(\cdot)$ and $GMP(\cdot)$ represent global average pooling and global maximum pooling respectively.
The global average pooling features are represented as yellow rectangles in Figure~\ref{fig:overview}, while the global maximum pooling features are represented as green rectangles.
The final video-level feature representation for source video is then calculated as:
\begin{eqnarray}
\label{eq:graph_layer}
\H_s = \sum_{l=1}^3{\mathbf{h}_s^l}
\end{eqnarray}
The ``$\oplus$'' symbol in Figure~\ref{fig:overview} means the summation operation.
The final video-level feature representation for target domain {$\H_t$} can be calculated similarly.
Therefore, the whole graph convolutional neural network can be formulated as:

\begin{eqnarray} 
\label{eq:gnn}
& GNN (\mathbf{V}_s,\A_s) \nonumber  \\
& =  \Psi_3(\Psi_2(\Psi_1(\mathbf{V}_s,\A_s))).
\end{eqnarray}


{
We use $G(\cdot)$ to represent the whole graph module for simplicity.
The node features $\mathbf{V}_s$, $\mathbf{V}_t$ are represented by $\Phi(\Z_s)$, $\Phi(\Z_t)$.
Further, combined with Eq.~\ref{eq:graph_layer} and Eq.~\ref{eq:gnn}, the video-level features can be formulated as:
}
\begin{eqnarray}
\H_s = G (\Phi(\Z_s)), 
\H_t = G (\Phi(\Z_t))
\end{eqnarray}

{
Finally, our overall optimization objective can be defined as follows:}
\begin{equation}
\begin{aligned}
    \theta_c^*, \theta_\Phi^*, \theta_G^* &= \mathop{\arg\min}_{\theta_c, \theta_\Phi, \theta_G}\ \ \mathcal{L}_c(C_s(\Phi(\mathcal{Z}_s)), \mathcal{Y}_s) \\
    &- \alpha \mathcal{L}_d(G(\Phi(\mathcal{Z}_s)), G(\Phi(\mathcal{Z}_t)))
    \label{eq:opt1}
\end{aligned}
\end{equation}
\begin{equation}
    \theta_d^* = \mathop{\arg\min}_{\theta_d}\ \ \mathcal{L}_c(C_s(\Phi(\mathcal{Z}_s)), \mathcal{Y}_s) + \alpha \mathcal{L}_d(G(\Phi(\mathcal{Z}_s)), G(\Phi(\mathcal{Z}_t)))
    \label{eq:opt2}
\end{equation}
{
where $\mathcal{L}_c$ is the classification loss, $\mathcal{L}_d$ is the domain alignment loss,
$\theta_c$ is the parameter of the source classifier $C_s(\cdot)$, 
$\theta_\Phi$ is the parameter of the mapping function $\Phi(\cdot)$, 
$\theta_\G$ is the parameter of the graph module $G(\cdot)$, 
$\theta_d$ represents the parameters of domain alignment discriminators.
More details about the domain alignment discriminators will be discussed in Section~\ref{sec:da}.
}
\subsection{Domain Alignment Backbone}
\label{sec:da}

We apply the above domain alignment loss function for two types of alignments: 

1) Domain alignment on the image/frame-level, which measures the image/frame feature similarity across the frames of source videos and target videos. The GNN node features are used as the frame-level features for this loss function.

2) Domain alignment on the video level, which measures video-level feature similarity between the source and target. We use {the output feature from of the GNN module} as the video feature representations.

We denote $D_f$ and $D_v$ as the two domain discriminators at frame-level and video-level, respectively. {$\mathcal{L}_{df}$ and $\mathcal{L}_{dv}$ are the corresponding losses to $D_f$ and $D_v$.}

In our framework, we use two SOTA DA backbones: Domain Adversarial Neural Network (DANN) and Conditional Adversarial Domain Adaptation (CDAN).
{
Either one can be applied to the overall in Eq.~\ref{eq:opt1} and Eq.~\ref{eq:opt2}}

{
We illustrate our strategy on these two standard adaptation frameworks purposely for the sake of generality on other modern UDA frameworks. 
Since most of the modern UDA frameworks utilize the Gradient Reversal Layer(GRL) to achieve the domain alignment, our model is capable of adopting those frameworks easily.
}

\subsubsection{Domain Adversarial Neural Network (DANN)}: DANN imposes adversarial learning to align the source and target features. 
It trains two discriminator $D_f, D_v$ to distinguish the source and target features at frame-level and video-level. 
Meanwhile, the feature extractor aims to generate domain-invariant features to fool the discriminator~\cite{ganin2015unsupervised}. 
The {DANN domain discrepancy loss} is defined as:
\begin{equation}
\begin{aligned}
&\mathcal{L}_{d} = 
\mathcal{L}_{d_f}+\mathcal{L}_{d_v},
\end{aligned}
\end{equation}
where 
{
\begin{equation}
\begin{aligned}
\mathcal{L}_{d_v}&=
-\mathbb{E}_{\Z_i^s \sim \mathcal{Z}_s}\log [D_v(G(\Phi(\Z_i^s))] \\
&- \mathbb{E}_{\Z_i^t \sim \mathcal{Z}_t}\log[1-D_v(G(\Phi(\Z_i^t)))],
\end{aligned}
\end{equation}
}
{
\begin{equation}
\begin{aligned}
\mathcal{L}_{d_f}& =
-\mathbb{E}_{\Z_i^s \sim \mathcal{Z}_s}\log [D_f(\Phi(\Z_i^s))]\\
&- \mathbb{E}_{\Z_i^t \sim \mathcal{Z}_t}\log[1-D_f(\Phi(\Z_i^t))],
\end{aligned}
\end{equation}
}
$\Phi$ is a two-layer image-level feature learning network and the frame-level loss is similar to the loss in conventional image-based domain adaptation in Equation~\ref{eq:da_loss1}. 

\subsubsection{Conditional Adversarial Domain Adaptation (CDAN)}: Similarly, CDAN adopts conditional adversarial learning to align the joint distribution of features and prediction of classifiers (CDAN)~\cite{long2018conditional}. 
 The CDAN loss function is:
\begin{equation}
 \begin{aligned}
	\mathcal{L}_{d} &  = \mathcal{L}_{d_f}+\mathcal{L}_{d_v}
	\end{aligned}
\end{equation}
where
{
\begin{equation}
\begin{aligned}
\mathcal{L}_{d_f} &=
	-\mathbb{E}_{\Z_i^s \sim \mathcal{Z}_s}
	 \log [D_f(\Phi(\Z_i^s)), C_s(\Phi(\Z_i^s))]\\
	& - \mathbb{E}_{\Z_i^t \sim \mathcal{Z}_t}\log[1-D_f(\Phi(\Z_i^t)), C_s(\Phi(\Z_i^t))]
	\end{aligned}
\end{equation}
}
{
\begin{equation}
\begin{aligned}
    	 \mathcal{L}_{d_v} &=
    	-\mathbb{E}_{\Z_i^s \sim \mathcal{Z}_s}
    	\log [D_v(G(\Phi(\Z_i^s))), C_s(G(\Phi(\Z_i^s)))] \\
    	&- \mathbb{E}_{\Z_i^t \sim \mathcal{Z}_t}\log[1-D_v(G(\Phi(\Z_i^t))), C_s(G(\Phi(\Z_i^t)))]
\end{aligned}
\end{equation}
}

\subsection{Our Framework}

Our proposed framework is shown in~Figure~\ref{fig:overview}.
For the video classifier, we compose it using a three-layer GNN and a three-layer MLP. 
The video-level feature is the summation of the three GNN layer features. 
Feature in each layer is obtained by concatenating the global average pooling feature and the global maximum pooling feature together.
The MLP is for the final classification task.
Both video-level domain discriminator and frame-level domain discriminator have a gradient reversal layer (GRL) inserted at the beginning of each module. 
{As shown in Figure~\ref{fig:overview}, there are three gradient backward-propagation paths. 
Two of them come from the domain discriminator losses $L_{df}$ and $L_{dv}$, which have the GRL. 
The GRL reverses the gradients before it to encourage the mapping function $\Phi$ to produce the domain-invariant representations.}
The implementation option of the following domain alignment backbone can be chosen from DANN and CDAN.

\section{Experiments}
\label{sec:exp}
\subsection{Datasets}
We use three popular video DA benchmark datasets in our experiments, which are {UCF-HMDB}$_{small}$~\cite{tang2016cross}, {UCF-HMDB}$_{full}$ \cite{videoDAICCV}, and {UCF-Olympic} \cite{jamal2018deep}.
The {UCF-HMDB}$_{small}$ and {UCF-HMDB}$_{full}$ datasets are two subsets extracted from UCF101~\cite{soomro2012ucf101} and HMDB51~\cite{6126543} with overlapped categories. 
The {UCF-HMDB}$_{small}$ has 5 shared classes and 1171 videos in total. 
The {UCF-HMDB}$_{full}$ is extended from {UCF-HMDB}$_{small}$ created by~\cite{videoDAICCV}, which has overall 12 overlapped categories and 3209 videos. 
The {UCF-Olympic} is the overlapped subset between UCF101~\cite{soomro2012ucf101} and Olympic Sports~\cite{niebles2010modeling} Dataset, which has 6 shared classes including Basketball, CleanAndJerk, ThrowDiscus, Diving, PoleVault, and TennisSwing, with a total of 1000 videos. \textbf{ Please refer to the supplementary material for more results. }

\subsection{Implementation details}

The experiments are performed on the Pytorch platform and using two Geforce RTX 3090 GPUs. 
Two different backbones are evaluated with our model:  1) Resnet-101 pretraiend on ImageNet~\cite{imagenet_dataset}; 2)  RGB I3D Network pretrained on ImageNet.
We use the features extracted from the source and target videos as the input of all competing methods. 
The frame number of all videos varies between 22 to 1184 in our dataset. 
In ~\cite{videoDAICCV}, they sampled all videos into 5 frames to reduce memory cost. 
{
Given that our approach is more spatially efficient, we set the number of sampled frames to be 40.
}
It is worth noting that the size of the sampling number can be changed based on video length. 
In our experiments, the Resnet-101 feature size is 2048 and the I3D feature size is 1024.
The visual similarity edges are constructed based on choosing top $K$ similar node features. 
We set the $K=5$ based on the validation dataset.
A more detailed ablation study on $K$ can be found in supplementary materials.
{
The video-level and frame-level domain discriminators $D_v$ and $D_f$ are both three-layer MLPs.}
We set up the batch size as 32 and use stochastic gradient descent (SGD) as the optimizer with a momentum of 0.9 and weight decay of $10^{-4}$. 
The learning rate is initialized to be 0.001. 
We trained our model for 100 epochs. 

\subsection{Benchmark Methods:}

We compare our method with several state-of-the-art approaches detailed as below:
\begin{itemize}
  
\item 1) \textit{Temporal Attentive Adversarial Adaptation Network (TA$^3$N)}~\cite{videoDAICCV}: a domain adaptation model using temporal attention and temporal pooling to align the temporal space between the source and target domains. 

\item  2)\textit{ Shuffle and Attend: Video domain Adaptation (SAVA)}~\cite{sava_eccv20}: a domain adaptation model focusing on extracting domain-invariant spatial features by randomly shuffling video frames and learning the visual features which are invariant to temporal orders on videos of both source and target domains. 
    
\item  3)\textit{ Temporal Co-attention Network (TCoN)}~\cite{Pan2020AdversarialCA}: a domain adaptation method with temporal co-attention designed on both source and target videos for activity recognition. 
This method extracts local temporal video segments and applies a temporal co-attention model on the source and target videos. 

All above three methods (1,2,3) extract sub-videos for each source or target video, thus, these methods require a large memory size. 
To reduce memory size, (1) samples their videos to be 5 frames, (2) samples their videos to be 16 frames regardless of the length of videos. 

\item 4) \textit{Adversarial Bipartite Graph  (ABG)}~\cite{ABG}: a video domain adaptation method that focuses on learning the frame-level bipartite graph between source and target videos frames. It does not extract a large number of sub-videos and does not use temporal attention. However, this approach first paired the source and target videos and used the paired videos as input. Pairing videos is NP-hard and can not scale up to large size video datasets.

\item  5) \textit{Spatio-temporal Contrastive Domain Adaptation (STCDA)}~\cite{song2021spatio}: a self-supervised contrastive framework for action recognition. Contrastive learning is introduced to learn domain-invariant video representations. 
The criticism is that a large number of negative sub-videos are needed to train the contrastive domain adaptation network. 
Storing these negative sub-videos is extremely memory expensive. 

\item 6) 
{
\textit{Learning Cross-Modal Contrastive Features for Video Domain Adaptation(LCMCF)}~\cite{kim2021learning}: a unified framework for video domain
adaptation using contrastive learning (CL) to learn cross-modal (RGB and flow) and cross-domain features.
The drawback of this method is the same as above 5), contrastive learning needs high computational complexity and spatial complexity.
}
\end{itemize}

\textbf{Metrics}: In our experiment, we use top-1 accuracy as the evaluation metric to compare our method with all four benchmark methods.
  
\subsection{Comparison with Benchmark Methods}
We show the quantitative comparison on {UCF-HMDB}$_{small}$, {UCF-HMDB}$_{full}$, and {UCF-Olympic} in Table~\ref{table:results}. 
Our method, using I3D features and CDAN backbone, performs best on almost all datasets compared to all baselines using I3D features.
As for features, since I3D features capture more temporal information via 3D convolution networks, it performs better than Resnet-101 features in most cases.

Besides, Conditional Adversarial Domain Adaptation(CDAN) has an overall higher accuracy than Domain Adversarial Neural Network (DANN).
It is worth noting that, by using only RGB I3D features, our method can achieve better performance than STCDA~\cite{song2021spatio}, which uses both RGB I3D features and flow I3D features.
This suggests that the proposed graph-based video domain adaptation framework has its own structural advantage.

{However, we can notice that our method using Resnet-101 features with CDAN backbone performed slightly worse than ABG in the alignment direction of H$\rightarrow$U under UCF-HMDB$_full$ dataset, even though we outperformed them in the alignment direction of U$\rightarrow$H.
This may because our method mainly focuses on temporal relations and lacks spatial processing techniques to focus on important spatial regions. 
ABG measures the similarity of source-target frame pairs for better spatial features but needs high memory ($O(n^2v^2)$ compared to ours $O(nv)$. Please refer to Table~\ref{tab:cost})
}

\begin{table*}[htp]
\caption{Quantitative results (\%) of our method compared with the benchmark methods (S$\rightarrow$T stands for the alignment from source to target data, U$\rightarrow$H represents using UCF as the source, and HMDB as the target. U$\rightarrow$ O represents using UCF as the source, Olympic as the target, and vice versa. (R) means taking only RGB modality. (R+F) means the backbone takes both RGB and Flow modalities together.)}
\centering
\begin{tabular}{l|l|p{0.11\linewidth}l|p{0.1\linewidth}l|p{0.1\linewidth}l}
\multirow{2}{*}{\begin{tabular}[c]{@{}l@{}}Method\\ \textit{(S$\rightarrow$T)}\end{tabular}} & \multirow{2}{*}{\parbox{1.5cm}{Feature encoder}} & \multicolumn{2}{c|}{{{UCF-HMDB}$_{small}$}} & \multicolumn{2}{c|}{{UCF-HMDB}$_{full}$} & \multicolumn{2}{c}{{UCF-Olympic}} \\ \cline{3-8} 
                    &   &\textit{U$\rightarrow$H}&\textit{H$\rightarrow$U}    &\textit{U$\rightarrow$H}   & \textit{H$\rightarrow$U}  & \textit{U$\rightarrow$O }    &\textit{O$\rightarrow$U}       \\ \hline
TA$^3$N              & Resnet-101& \textbf{99.33}             & 99.47         & 78.33         & 81.79          & \textbf{98.15}         & 92.92        \\ \hline
SAVA        &I3D(R)& -         & -             & 82.2            & 91.2         & -          & -        \\ \hline
TCoN        &BNIncept(R)& -        & -              & \textbf{87.24}       & 89.06          & 96.82          & 96.79        \\ \hline
ABG         & Resnet-101  & \textbf{99.33}  & 98.41     & 79.17         & 85.11          & \textbf{98.15}          & 92.50        \\ \hline
STCDA       & \parbox{2.2cm}{BNIncep(R+F)} &98.7 & \textbf{100.0} & 80.0 & 87.7 & 98.1 & 96.3 \\ \hline
STCDA       & I3D(R+F) & -&-& 83.1& 92.1&-&- \\ \hline
LCMCF       & I3D(R+F) & -&-& 84.7 & 92.8 &-&- \\ \hline
\textbf{Ours-DANN }  &Resnet-101& 98.66  & 97.88 & 74.72 & 78.63 & 94.44 &90.41\\
\textbf{Ours-DANN }  &I3D(R)& 98.66  & 98.94 & 82.22 & 91.59 & 96.29 & 95.83 \\ 
\textbf{Ours-CDAN }  &Resnet-101&  \textbf{99.33}       & 99.47     &  79.72 & 83.18  & \textbf{98.15} & 92.5 \\ \textbf{Ours-CDAN }  &I3D(R)&   \textbf{99.33}  & \textbf{100.0} &  85.83  &  \textbf{93.87}  & \textbf{98.15}  &  \textbf{97.91}  \\ 
\end{tabular} 
\label{table:results} 
\end{table*}
 
\subsection{Ablation Study and Analysis}
\subsubsection{Visual similarity edge and frame number}
We designed an ablation study on the graph visual similarity edge with the different number of video frames in Table~\ref{table:frame}. 
The experiment is conducted  on {UCF-HMDB}$_{full}$ dataset. 
  In general, more frames bring higher accuracy. Therefore, it is crucial to handle more frames efficiently with limited GPU memory. Although we have the capability to handle more frames, we set the frames number to be 40 in our experiments to speed up the training process.
  
It is also shown that the performance consistently increases when including the visual similarity edge in our graph model.
In addition, the ablation study in Table~\ref{table:simi} also demonstrated the performance improvement by using visual similarity edged in our graph model.

\subsubsection{Visual Similarity Edge} 
We perform an ablation study on the visual similarity edge in Table~\ref{table:simi} using Resnest101 features.
$D_1$, $D_2$ and $D_3$ represent {UCF-HMDB}$_{small}$, {UCF-HMDB}$_{full}$ and {UCF-Olympic} respectively. 
We show the performance of our model with two settings: 1) with only temporal relation edges in the graph; 2) with both temporal relation edges and visual similarity edges in the graph.
We can see that our method with both temporal relation and similarity edges has a consistently better performance than temporal edge only baseline.

\begin{table}[htp]
\setlength\tabcolsep{4pt}
\caption{Ablation study of different frame numbers and similarity edges on {UCF-HMDB}$_{full}$ dataset. w represents with, w/o represents without.}
\centering
\begin{tabular}{l|c|ll|ll}
                                           &              & \multicolumn{2}{c|}{Resnet-101}       & \multicolumn{2}{c}{I3D}               \\ \cline{3-6} 
                                           & {frame \#} & {U$\rightarrow$H} & {H$\rightarrow$U} & {U$\rightarrow$H} & {H$\rightarrow$U} \\ \hline
\multirow{4}{*}{ \parbox{3cm}{w/ temporal but w/o similarity edge}} & 16           & 77.22             & 80.03             & \textbf{83.05}    & \textbf{92.11}    \\
                                           & 24           & {77.77}    & {79.50}      & {81.66}             & {91.41}             \\
                                           & 32           & {78.83}             & {80.03}             & {82.77}             & {91.06}             \\
                                           & 40           & {\textbf{79.17}}    & {\textbf{82.31}}    & {\textbf{83.05}}    & {91.94}             \\ \hline
\multirow{4}{*}{ \parbox{3cm}{w/ temporal and similarity edge}}  & 16           & {75.5}              & {80.91}             & 85.27             & 89.66             \\
                                           & 24           & {76.38}             & {\textbf{83.18}}    & {85.55}             & {91.59}             \\
                                           & 32           & {78.05}             & {82.66}             & {82.77}             & {93.52}    \\
                                           & 40           & {\textbf{79.72}}    & {83.01}             & {\textbf{85.83}}    & {\textbf{93.87}}            
\end{tabular}
\label{table:frame}
\end{table}

\subsubsection{Optimal number of the similarity edges }
To investigate the optimal selection of numbers of similarity edges,
we designed the ablation study with different top-K values as shown in Table~\ref{table:topk}.
The experiment was conducted on the UCF-HMDB$_{full}$ dataset with a sample frames number of 40.
U$\rightarrow$H represents using UCF101 as the source and HMDB51 as the target. 
H$\rightarrow$U represents using HMDB51 as the source and UCF101 as the target.
We tested the performance on both Resnet-101\cite{he2016deep} and RGB-based I3D~\cite{carreira2017quo} features.
The results show that fewer number of similarity edges has a better performance. 
However, when there are no similarity edges, the performance is poor.
The best performance was achieved with a similarity edge number of 5. 
After the number of 5, accuracy decreases as the number of similarity edges increases. 
Although in the alignment of H$\rightarrow$U, selecting a higher number of similarity edges using I3D features also gave good accuracy, the setting of K=5 still is still the preferred selection.

\begin{table}[ht]
\setlength{\tabcolsep}{4pt}
\centering
\caption{Ablation study of the number of the similarity edges based on the UCF-HMDB$_{full}$~\cite{videoDAICCV} dataset. U$\rightarrow$H denotes using UCF101~\cite{soomro2012ucf101} as the source and HMDB51~\cite{6126543} as the target. H$\rightarrow$U denotes using HMDB51 as the source and UCF101 as the target.  }
\begin{tabular}{c|c|c|c|c}
\multirow{2}{*}{Top-K value} & \multicolumn{2}{c|}{\textit{ U $\rightarrow$ H }}       & \multicolumn{2}{c}{\textit{ H $\rightarrow$ U }}\\ \cline{2-5} 
                   & \multicolumn{1}{l|}{Resnet} & \multicolumn{1}{l|}{I3D} & \multicolumn{1}{l|}{Resnet}   & I3D   \\ \hline
0                 &    79.17     & 82.31 &   83.05    & 91.94 \\ \hline
5                  &  \textbf{79.72}            &    \textbf{85.83}     &    \textbf{83.18}   & \textbf{93.87}   \\ \hline
10                 &  77.22           &      85.0    &     80.21             &   82.83   \\ \hline
15                 &  73.88               &  83.88           &    79.85          &  88.96 \\ \hline
20                 &  75.0                &    83.61        &   80.21         & 85.81\\ \hline
25                 &  75.88                 &  84.72               &      77.93              &    93.34
\end{tabular}
\label{table:topk}
\end{table}

\subsubsection{Domain discriminators} Our method contains two-level domain discriminators, the frame-level domain discriminator $D_f$ and the video-level domain discriminator $D_v$.
In Table~\ref{table:gain}, We show the comparisons between $D_f$ and $D_v$ using Resnet-101 features on {UCF-HMDB}$_{full}$ dataset. 
{
As demonstrated, both $D_f$ and $D_v$ played a role in aligning source and target.
$D_f$ obtained 2.23\% and 2.27\% gains in U$\rightarrow$H and H$\rightarrow$U, respectively.
Nevertheless, $D_v$ is observed to achieve higher performance gains of 6.12\% in U$\rightarrow$H and 6.83\% in H$\rightarrow$U.
Moreover, the results of $D_f$+$D_v$ display a significant improvement in U$\rightarrow$H compared to $D_f$ only and $D_v$ only.
Although $D_f+D_v$ shows slightly lower performance than $D_v$ in the H$\rightarrow$U, the averaged performance in two directions is improved. 
This experiment suggests it is useful to include both video-level and frame-level discriminators in our graph-based video DA framework.  
}


\begin{table}[htp]
\setlength\tabcolsep{4pt}
\caption{Comparison of different domain discriminators on {UCF-HMDB}$_{full}$ using Resnest-101 features. $D_f$ represents the frame-level domain discriminator. $D_v$ represents the video-level domain discriminator.}
\centering
\begin{tabular}{c|ll|ll}
            & \multicolumn{2}{c|}{U$\rightarrow$ H} & \multicolumn{2}{c}{H$\rightarrow$ U} \\ \cline{2-5} 
            & \multicolumn{1}{l|}{acc.}     & gain   & \multicolumn{1}{l|}{acc.}    & gain   \\ \hline
source only & \multicolumn{1}{l|}{71.38}    & -      & \multicolumn{1}{l|}{76.53}   & -      \\ \hline
target only & \multicolumn{1}{l|}{85.85}    & -      & \multicolumn{1}{l|}{94.92}   & -      \\ 
$D_f$        & \multicolumn{1}{l|}{73.61}    & 2.23 & \multicolumn{1}{l|}{78.8}   &  2.27   \\ 
$D_v$         & \multicolumn{1}{l|}{77.50}  & 6.12   & \multicolumn{1}{l|}{\textbf{83.36}}   & \textbf{6.83}  \\ \hline
$D_f$+$D_v$      & \multicolumn{1}{l|}{\textbf{79.72}}    & \textbf{8.34}   & \multicolumn{1}{l|}{ 83.18}   & 6.65  
\end{tabular}
\label{table:gain}
\end{table}

\begin{table}[ht]
\setlength\tabcolsep{4pt}
\caption{Ablation Study on similarity edges(\%) on our proposed graph video DA method. w represents with, w/o represents without.}
\centering
\begin{tabular}{p{3cm}|ll|ll|ll}
\multirow{2}{*}{{Ablation Design}} & \multicolumn{2}{c|}{{$D_1$}} & \multicolumn{2}{c|}{$D_2$} & \multicolumn{2}{c}{$D_3$} \\ \cline{2-7} 
                                 & \parbox{.7cm}{U$\rightarrow$H} & \parbox{.7cm}{H$\rightarrow$U}       & \parbox{.7cm}{U$\rightarrow$H}       & \parbox{.7cm}{H$\rightarrow$U}     & \parbox{.7cm}{U$\rightarrow$O}    &\parbox{.7cm}{O$\rightarrow$U}   \\ \hline

{w/ temporal edge but w/o similarity edge} &    {\textbf{99.33}}  &  {98.94}  &   {82.31}     &  {83.05}   &  {\textbf{98.15}}   & {\textbf{96.29}} \\ \hline
{w/ temporal and similarity edge}  &   {\textbf{99.33}}  & {\textbf{99.47}}  & {\textbf{83.18}} & {\textbf{85.83}} &  {\textbf{98.15}}  &  {\textbf{96.29}}   \\ 
\end{tabular}
\label{table:simi}
\end{table}

\subsection{t-SNE visualization}
\begin{figure}[ht]
  \centering
  \begin{subfigure}{0.45\linewidth}
    \centering
    \includegraphics[scale = 0.25]{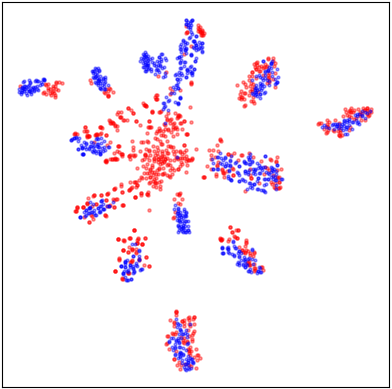}
    \caption{Source only}
    \label{fig:tsne-a} 
  \end{subfigure}
  \hfill
  \begin{subfigure}{0.45\linewidth}
    \centering
    \includegraphics[scale = 0.25]{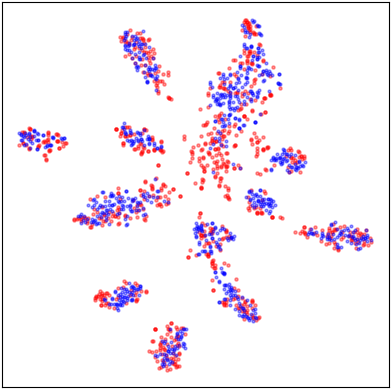}
    \caption{Our method with $D_f$ only}
    \label{fig:tsne-b}
  \end{subfigure}
  
  \begin{subfigure}{0.45\linewidth}
    \centering
    \includegraphics[scale = 0.25]{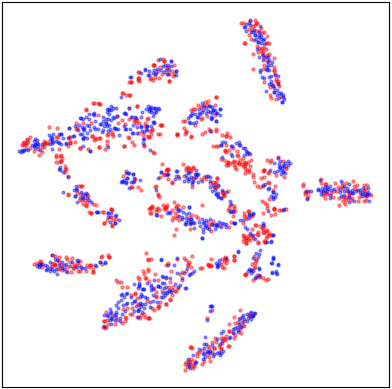}
    \caption{Our method with $D_v$ only}
    \label{fig:tsne-c}
  \end{subfigure}
  \hfill
  \begin{subfigure}{0.45\linewidth}
    \centering
    \includegraphics[scale = 0.25]{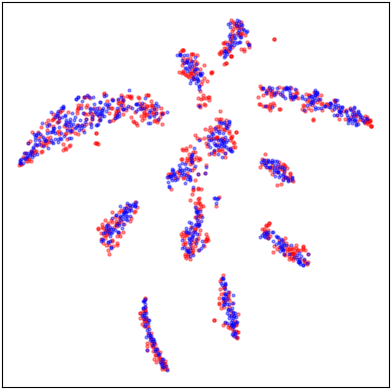}
    \caption{Our method with both $D_f$ and $D_v$.}
    \label{fig:tsne-d}
  \end{subfigure}
  \caption{The comparison of t-SNE visualization. The blue points represent the source data(UCF101), while the red points represents the target data(HMDB51). }
  \label{fig:tsne}
\end{figure}

Additionally, the t-SNE\cite{maaten2008visualizing} visualizations of the distribution of the source and target with different settings are shown in Figure~\ref{fig:tsne}. Our full model with both video-level domain loss $D_v$ and frame-level domain loss $D_f$ properly aligns the distribution of the target with the source domain. Figure~\ref{fig:tsne-a} shows that the source-only model poorly aligns the distributions between source and target. Figure~\ref{fig:tsne-b} shows that our method with frame-level domain loss helps align the source and target distributions, but some target data(red points) still scatter between clusters. In Figure~\ref{fig:tsne-c}, our method with video-level domain loss reduces the number of data points scattered between the clusters, but some clusters are too close to each other. Our method with both video-level domain loss and frame-level domain loss successfully aligns the distribution of two domains. 

\subsection{Comparison of GPU memory usage.}
\subsubsection{Quantitative comparison}
{
We compare the GPU memory usage of our method with two of the benchmark methods that provided the code, TA$^3$N~\cite{videoDAICCV} and ABG~\cite{ABG}. 
}
To ensure a valid comparison experiment, all the methods used the same control hyper-parameters such as batch size and feature dimension. 
{All other settings remained the same as their default.}
{
Originally, we set the batch size to 32 and all hidden feature dimensions to be 2048.
However, ABG consumed too much memory usage and reached the GPU maximum at 10 frames.
In order to obtain more data, we reduce their hidden feature dimensions to 512. 
}
Then we alter the number of frames from 5 to 40 to observe how the GPU memory usage changes. All the experiments were conducted using one RTX 3090 GPU. 

{
The comparison results are shown in Figure~\ref{fig:fig1b}. 
The x-axis displays the number of frames, while the y-axis displays the GPU usage. 
The orange solid line is the trend of TA$^3$N's GPU memory usage as the number of frames increases. 
The orange dotted line represents the estimated tendency of the TA$^3$N's GPU memory usage. 
Similarly, The green solid line and dotted line are corresponding to ABG's GPU memory usage and the estimated tendency.
The blue line represents the trend of our method's GPU memory usage. 
The red dash line marks the maximum of the GPU memory.
}

In the Figure~\ref{fig:fig1b}, even though our method started at a higher GPU usage point when the frame number is 5, it grew very slowly. 
In the contrast, TA$^3$N started at a more memory-saving point, nevertheless, it rose very fast. 
Although there seems to be enough GPU memory available at the point of 30 frames, TA$^3$N failed to run properly due to the high computational cost. 
We estimated the TA$^3$N's later tendency based on the previous one from 5 to 25 frames point. 
As shown in the results. We reduced about $65\%$ memory cost at frame number 25 compared to TA$^3$N.

{
Regarding the ABG, it consumed more GPU usage, because ABG is required to generate paired source and target videos, which demands a high spatial cost of $O(n^2v^2)$ (refer to Table~\ref{tab:cost}). 
However, pairing is not required in our case. 
As shown in Figure~\ref{fig:fig1b}, although ABG was set to a lower hidden feature dimension, it still requested much greater GPU memory than others.
}

\subsubsection{Theoretical comparison}
{
With regard to the other benchmark methods' codes, we couldn't conduct the quantitative comparison with them since their codes were unavailable. 
Instead, we show the theoretical big O annotation comparisons in Table~\ref{tab:cost}, where n represents the number of sampled frames, m is the number of temporal sub-videos, v is the number of videos in a batch.
As we can observe that all the benchmark methods, except for ours, contain at least one squared term.
Because of the ability to extract the shared temporal relation in the latent space, our method doesn't need to deal with the spatial cost of frame/video pairs.
Hence, we don't have a squared term in our analyzed space complexity. 
}

\begin{table}[ht]
\begin{center}
\setlength\tabcolsep{4pt}
\caption{Theoretical analysis of space complexity comparison. n - the number of sampled frames; m - the number of temporal sub-videos; v - the number of videos in a batch}
\begin{tabular}{l|lllll|l}
\hline\noalign{\smallskip}
 Method & SAVA    & ABG     & STCDA   & {TCoN}    & {LCMCF}  & Ours  \\ \noalign{\smallskip}\hline\noalign{\smallskip}
\multicolumn{1}{c|}{\begin{tabular}[c]{@{}c@{}}Space\\ complexity\end{tabular}} & {$O(nm^2v)$} & {$O(n^2v^2)$} & {$O(nm^2v)$} & {$O(nm^2v)$} & {$O(nv^2)$} & {$O(nv)$} \\ \noalign{\smallskip}\hline
\end{tabular}
\label{tab:cost}
\end{center}
\end{table}

\section{Conclusion}
We proposed a computational efficient unsupervised video domain adaptation framework by modeling the videos in a graph model with attention convolution networks.
We compared our model to state-of-the-art video domain adaptation methods and achieve consistent improvement on cross-domain video classification tasks with significantly less computational memory cost.
Our method is memory efficient that is able to process long videos with large frame number. 
Although our method achieved high performance of domain alignment, only three datasets were tested. We expect to apply our method to more challenging datasets and solve real-world problems in the future.


%


\ifCLASSOPTIONcaptionsoff
  \newpage
\fi



%

\bibliographystyle{IEEEtran}
\bibliography{main}

%

\begin{IEEEbiography}[{\includegraphics[width=1in,height=1.25in,clip,keepaspectratio]{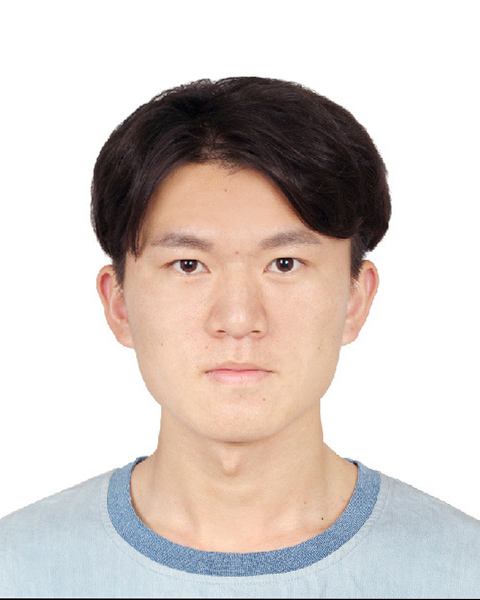}}]{Xinyue Hu}
received the bachelor’s degree in Artificial Intelligence from the University of Science and Technology, Beijing, in 2019. He is currently a phd student at the University of Texas at Arlington. His research interests include Computer vision, Machine Learning, Medical Imaging Analysis.
\end{IEEEbiography}
\begin{IEEEbiography}[{\includegraphics[width=1in,height=1.25in,clip,keepaspectratio]{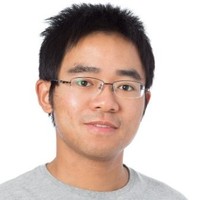}}]{Lin Gu}
received the B.Eng. degree from Shanghai University, Shanghai, China in 2009, and Ph.D. degree at the Australian National University and NICTA (Now Data61) in 2014. Currently, he is a research scientist at RIKEN AIP, Japan and a special researcher at the University of Tokyo. His research interests include image spectroscopy, shape recovery, colour science, and pattern recognition.
\end{IEEEbiography}
\begin{IEEEbiography}[{\includegraphics[width=1in,height=1.25in,clip,keepaspectratio]{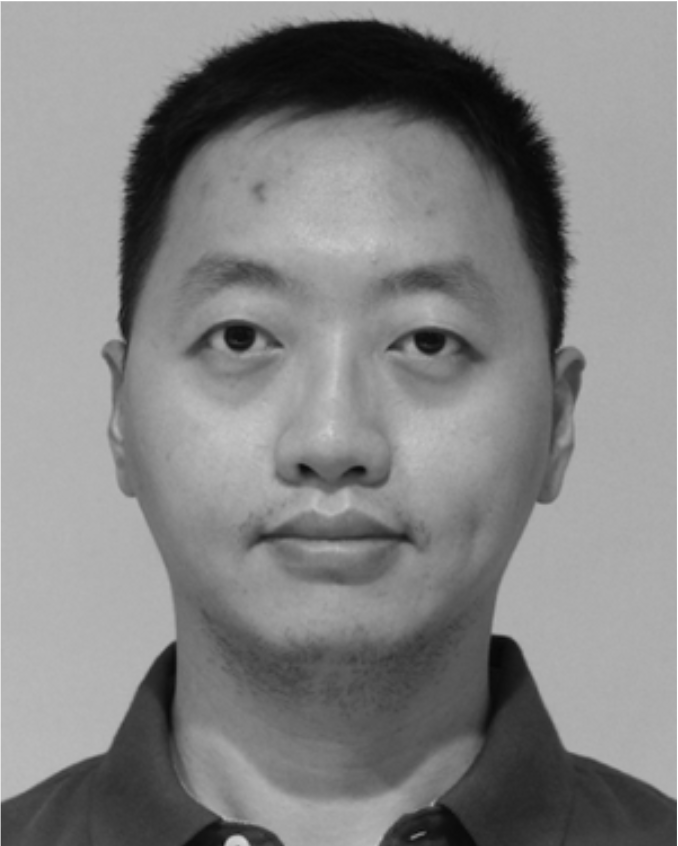}}]{Liangchen Liu}
received the B.Eng. degree in information engineering and the M.Sc. degree in instrument science and technology from Chongqing University, Chongqing, China, in 2009 and 2012, respectively, and the Ph.D. degree from The University of Queensland, Brisbane, QLD, Australia, in 2017. He is currently a Research Fellow with Clinical Center, National Institutes of Health. His  research interests include unsupervised learning, detection, segmentation, and visual attribute and its related applications.

\end{IEEEbiography}
\begin{IEEEbiography}[{\includegraphics[width=1in,height=1.25in,clip,keepaspectratio]{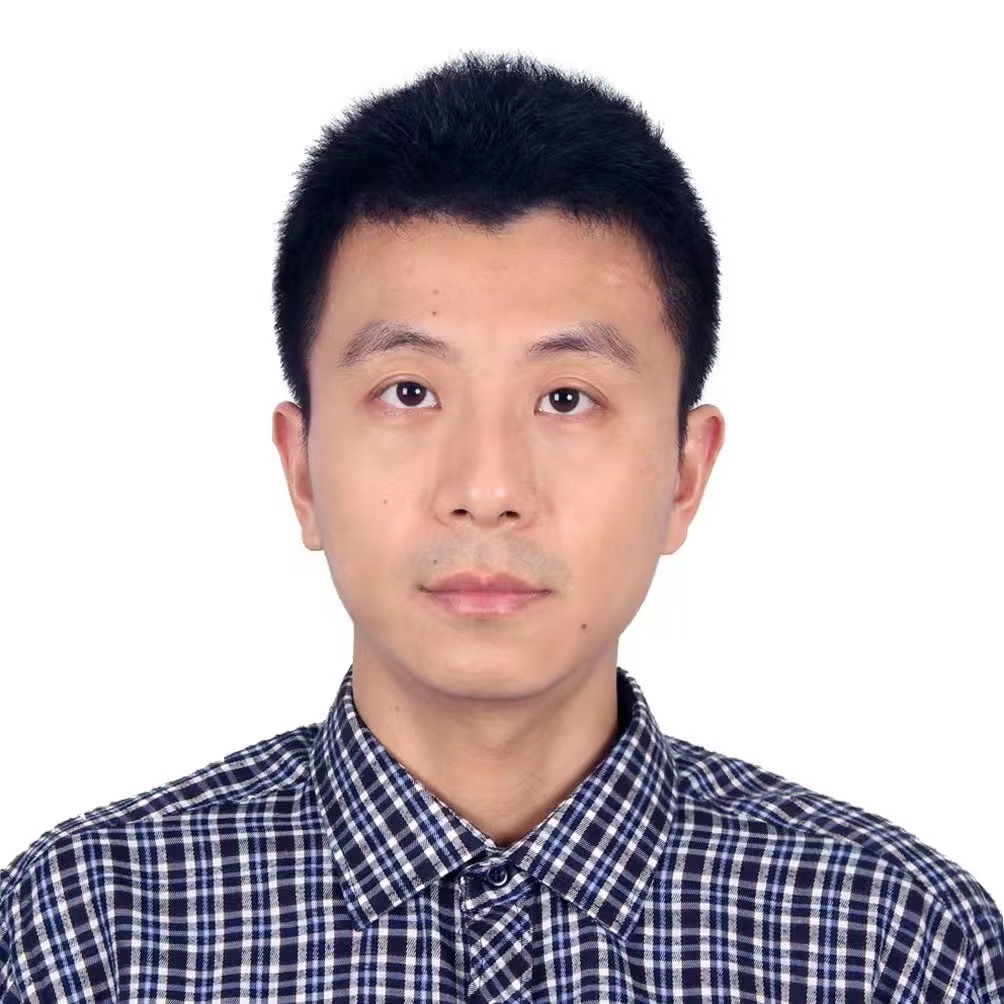}}]{Ruijiang Li}
received the PhD degree from Fudan University, Shanghai, China in 2013. He is a senior data science leader at Sanofi China. His current research interests include knowledge graph representation, graph generative models, domain adaptation and discrete operation.
\end{IEEEbiography}
\begin{IEEEbiography}[{\includegraphics[width=1in,height=1.25in,clip,keepaspectratio]{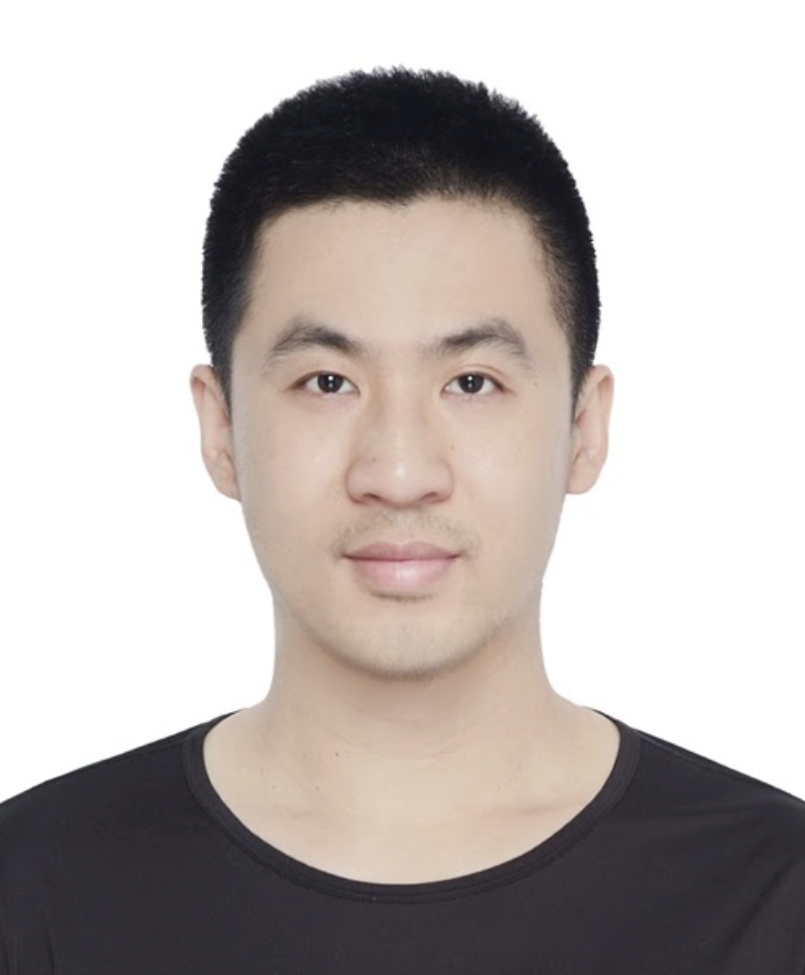}}]{Chang Su}
received his PhD in Control Science and Engineering from Xi'an Jiaotong university and completed his Postdoctoral fellowship in the Department of Population Health Sciences, Weill Cornell Medical College, Cornell University. He is currently an Assistant Professor in the Department of Health Service Administration and Policy (HSAP), College of Public Health, Temple University. He is broadly interested in machine learning, healthcare data mining, health informatics.
\end{IEEEbiography}
\begin{IEEEbiography}[{\includegraphics[width=1in,height=1.25in,clip,keepaspectratio]{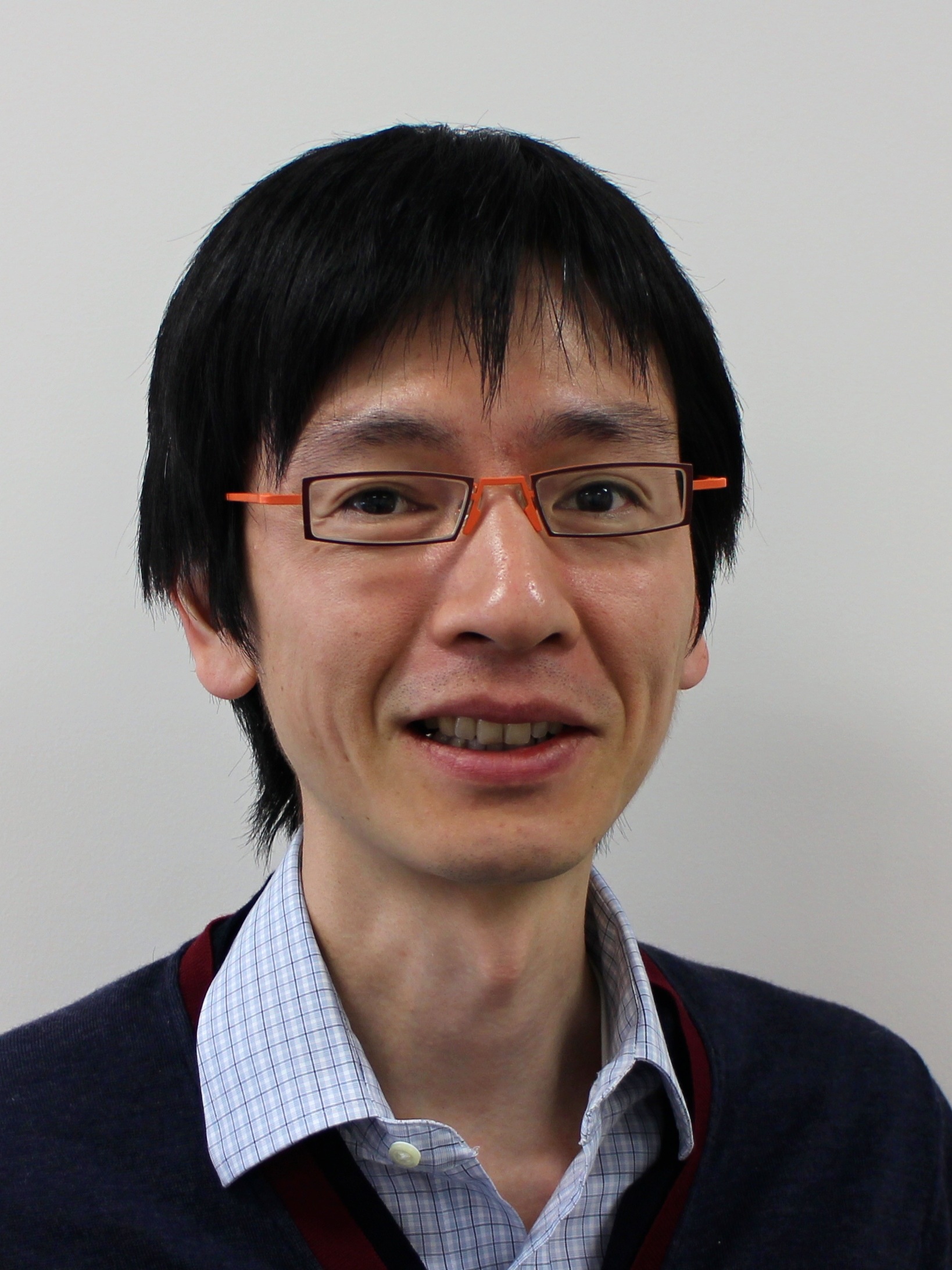}}]{Tatsuya Harada}
(Member, IEEE) received the Ph.D. degree in mechanical engineering from The University of Tokyo in 2001. He is currently a Professor with the Research Center for Advanced Science and Technology, The University of Tokyo; a Team Leader at the RIKEN Center for Advanced Intelligence Project (AIP), and a Visiting Professor at the National Institute of Informatics (NII). His research interests include visual recognition, machine learning, and intelligent robot.
\end{IEEEbiography}
\begin{IEEEbiography}[{\includegraphics[width=1in,height=1.25in,clip,keepaspectratio]{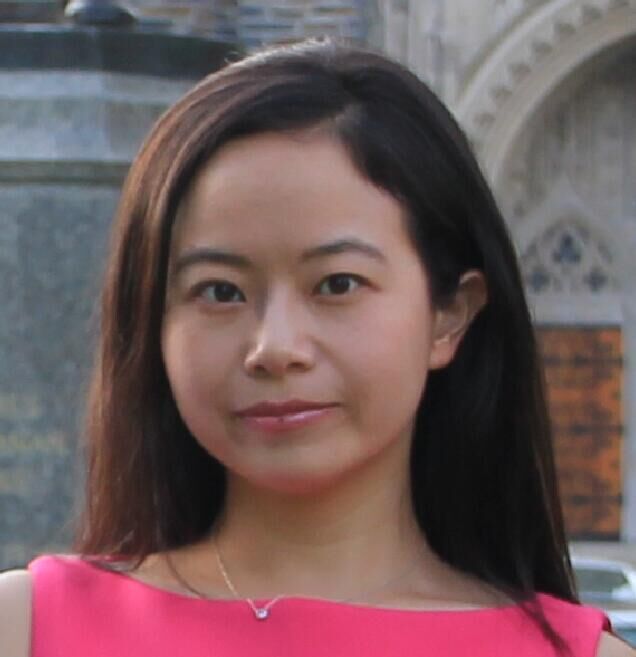}}]{Yingying Zhu}
received her Ph.D. degree from University of Queensland, Australia. She did a postdoc at Cornell University and a postdoc in UNC Chapel Hill. Currently she is working in the Computer Science and Engineering Department, University of Texas at Arlington as an assistant professor and also a guest researcher working in clinical center, NIH. She was a Staff Scientist working at Clinical Center, National Institutes of Health. Her research interests lie at the intersection of computer vision, medical image analysis, bioinformatics and machine learning.
\end{IEEEbiography}





\end{document}